\documentclass[10pt,twocolumn,letterpaper]{article}

\usepackage{iccv}
\usepackage{times}
\usepackage{epsfig}
\usepackage{graphicx}
\usepackage{amsmath}
\usepackage{amssymb}
\usepackage{booktabs}
\usepackage[table,xcdraw]{xcolor}
\usepackage{multirow}
\usepackage{pifont}
\usepackage{subcaption}
\usepackage{float}
\usepackage{amsmath}
\usepackage[font={small}]{caption}

\usepackage[pagebackref=true,breaklinks=true,letterpaper=true,colorlinks,bookmarks=false]{hyperref}

\iccvfinalcopy % *** Uncomment this line for the final submission

\def\iccvPaperID{5} % *** Enter the ICCV Paper ID here

\ificcvfinal\fi

\begin{document}

%%%%%%%%% TITLE
\title{Transformer-based Detection of Microorganisms \\ on High-Resolution Petri Dish Images}

% \author{Nikolas Ebert\\
% \and

\author{
Nikolas Ebert$^{1,2}$ \hfill
Didier Stricker$^{2}$ \hfill
Oliver Wasenm\"uller$^{1}$ 
\\
$^{1}$Mannheim University for Applied Science, Germany\\
$^{2}$RPTU Kaiserslautern-Landau, Germany\\
\tt\small 
n.ebert@hs-mannheim.de, 
didier.stricker@dfki.de, 
o.wasenmueller@hs-mannheim.de
}

\maketitle
\ificcvfinal\thispagestyle{empty}\fi

%%%%%%%%% ABSTRACT
\begin{abstract}
   Many medical or pharmaceutical processes have strict guidelines regarding continuous hygiene monitoring. This often involves the labor-intensive task of manually counting microorganisms in Petri dishes by trained personnel. Automation attempts often struggle due to major challenges: significant scaling differences, low separation, low contrast, etc. To address these challenges, we introduce AttnPAFPN, a high-resolution detection pipeline that leverages a novel transformer variation, the efficient-global self-attention mechanism. Our streamlined approach can be easily integrated in almost any multi-scale object detection pipeline. In a comprehensive evaluation on the publicly available AGAR dataset, we demonstrate the superior accuracy of our network over the current state-of-the-art. In order to demonstrate the task-independent performance of our approach, we perform further experiments on COCO and LIVECell datasets.
\end{abstract}

%%%%%%%%% BODY TEXT
\section{Introduction}

\begin{figure}[t]
    \centering
     \begin{subfigure}[b]{0.49\textwidth}
         \centering
         \includegraphics[width=.95\textwidth]{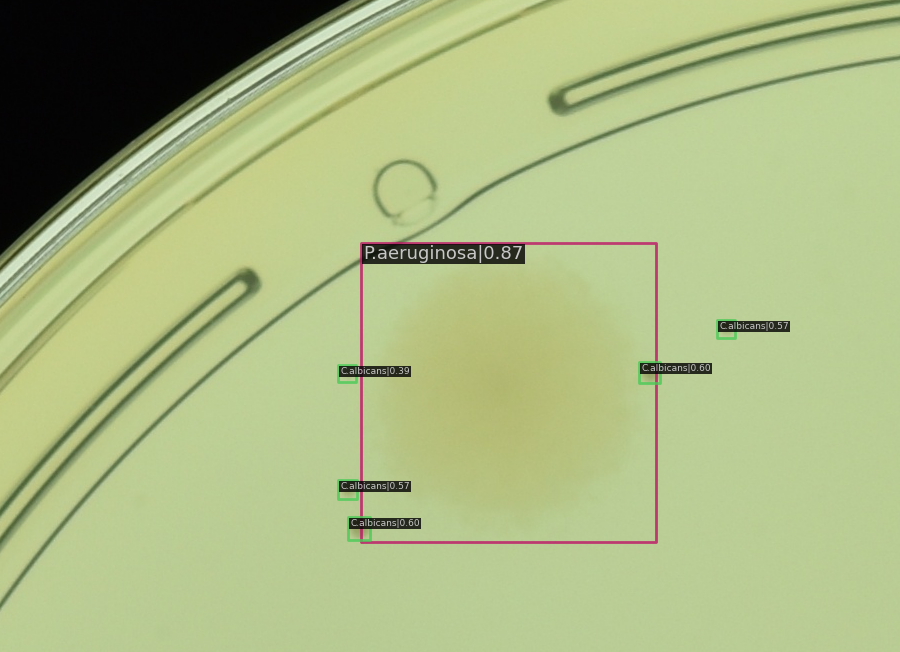}
     \end{subfigure}
     \\
     \begin{subfigure}[b]{0.49\textwidth}
        \centering
         \includegraphics[width=.95\textwidth]{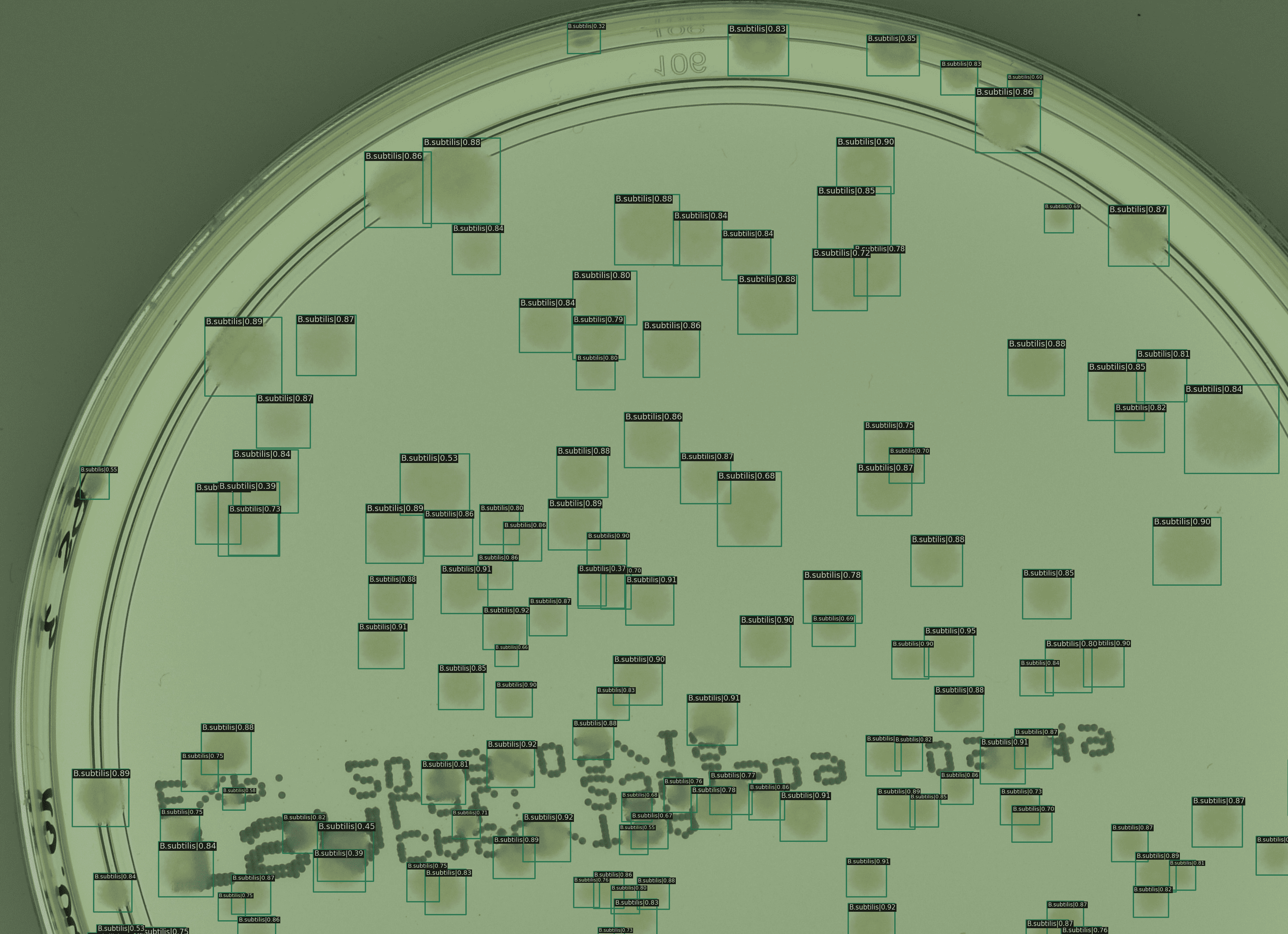}
     \end{subfigure}
     % \\
     % \begin{subfigure}[b]{0.49\textwidth}
     %     \centering
     %     \includegraphics[width=.8\textwidth]{images/SKOV3_Phase_E4_1_00d08h00m_2.png}
     % \end{subfigure}
\caption{The biggest challenges in hygiene monitoring are the detection of particularly small organisms, the significant variation in colony size, low contrast between foreground and background, as well as a high number of colonies with large overlap. The images show typical inference results of our method on the test data.}
    \label{fig:cover}
\end{figure}

Regulatory bodies such as the European Medicines Agency (EMA) and the U.S. Food and Drug Administration (FDA) mandate strict guidelines for continuous hygiene monitoring in the pharmaceutical, cosmetics and food industries.
As a result, a large number of Petri dishes must be examined for microbial colonies on a daily basis by experienced biologists, which is time-consuming and error-prone.
Automating this process presents several challenges. 
One is the high resolution required to reliably detect tiny colonies. Another is that colonies vary widely in size and shape and can overlap, making automated detection difficult (see Figure \ref{fig:cover}).
There are several open-source approaches \cite{geissmann2013opencfu,lamprecht2007cellprofiler,torelli2018autocellseg} that use classical computer vision techniques such as image filters and intensity variations to differentiate colonies from the agar-medium.
However, these processes are based on hand-crafted features and laborious to use.
%One of the major challenges in detecting microorganisms in Petri dished is the high resolution required to reliably detect tiny colonies.
%In addition, colonies differ greatly in size, shape, and can overlap, making automated detection difficult (see Figure \ref{fig:vergleich}).

Colony detection can be automated through the use of neural networks, such as Faster-RCNN \cite{majchrowska2021agar}, which have proven to be more accurate and robust than traditional computer vision methods.
Recently, transformer networks \cite{vaswani2017attention} were introduced, outperforming their convolutional-counterparts in most tasks \cite{liu2021swin,zhu2020deformable}.
This success is partly due to the self-attention mechanism, which enables transformers to model information spatial dependencies within large receptive fields.
%However, high-resolution image processing for hygiene monitoring requires a more efficient alternative to conventional SA, since it is much too computationally inefficient.
A drawback of standard self-attention is its quadratic complexity, resulting in large memory requirements and computational costs, especially when applied to high-resolution images for hygiene monitoring.

In this paper, we present an innovative approach to colony detection in the field of computer vision. 
Our method, called AttnPAFPN, leverages a novel efficient-global self-attention mechanism to improve the performance of a path aggregation feature pyramid network (PAFPN) \cite{liu2018path} for object detection.
In combination with further optimizations, our efficient-global self-attention achieves superior accuracy and performance, especially when processing high-resolution images.
Furthermore, we introduce new high-resolution prediction-heads to improve the detection of tiny objects.
A hallmark of our AttnPAFPN is its flexibility, as it can be integrated into almost any top-down object detection method.
To demonstrate this flexibility, we integrate our method into two general object detectors \cite{feng2021tood,Ren_2017,he2017mask}.
Augmented with our AttnPAFPN, these networks show superior performance in terms of accuracy over the current SoTA on the AGAR dataset \cite{majchrowska2021agar} for colony detection. In addition, we include an extensive ablation study of our method with varying image resolutions.
To demonstrate the task-independent performance of our approach, we also conduct experiments on COCO \cite{lin2014microsoft} for general object detection and on LIVECell \cite{edlund2021livecell} for the segmentation of cells in microscope images.

%%%%%%%%%%%%%%%%%%%%%%%%%%%%%%%%%%%%%%%%%%%%%%%%%%%%%%%%%%%%%%%%%%
\section{Related Works}\label{sec:rw}

\begin{figure*}[t]
\centering
\includegraphics[width=0.85\linewidth]{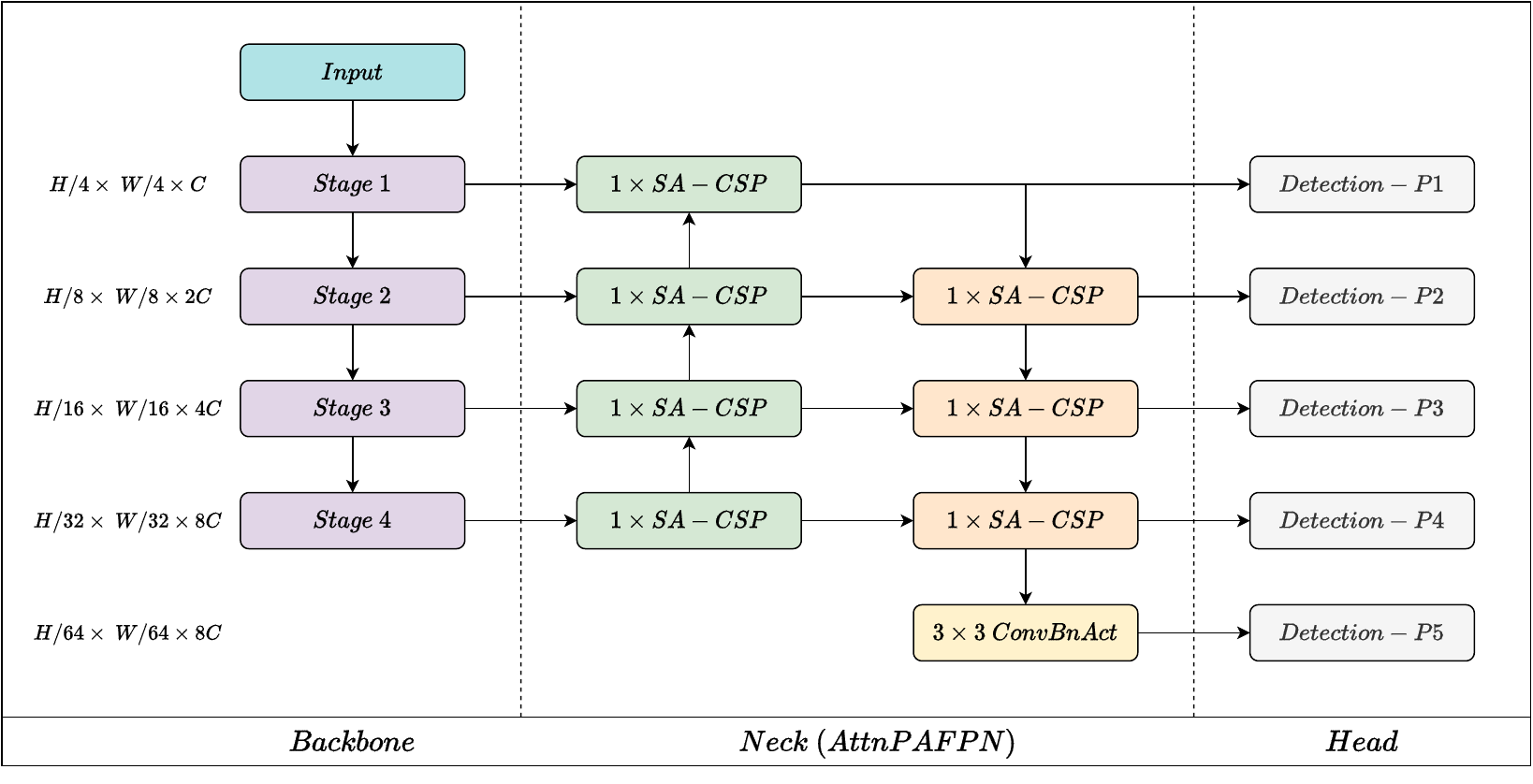}
\caption{\textbf{Architecture overview.} Our object detection network consists of a backbone network, a neck and a prediction-head. We use our AttnPAFPN as the neck, which consists of self-attention extended CSP-Bottlenecks (SA-CSP). Almost any method can be used for the final prediction by the head (e.g. TOOD \cite{feng2021tood}).}
\label{fig:architecture}
\end{figure*}

\subsection{Detecting colonies}
Automated colony counting has been of interest since the late 1950s \cite{alexander1958automatic,mansberg1957automatic}. 
Nowadays, there are several tools available, such as OpenCFU \cite{geissmann2013opencfu} and AutoCellSeg \cite{torelli2018autocellseg}, which assist in the detection of microorganisms, based on conventional computer vision methods. 
The main drawback of these tools is their limited automation, requiring handcrafted features for colony detection. 
Setting these features requires expert knowledge, similar to manual counting.

In addition to these conventional methods, several deep learning-based approaches \cite{falk2019u,ferrari2017bacterial,majchrowska2021agar,gorokhov2022bacterial,shamash2021onepetri} have been proposed for detecting colonies of microorganisms on agar plates.
Ferrari et al. \cite{ferrari2017bacterial} utilize convolutional neural networks (CNNs) for bacterial classification, resulting in significant improvements compared to handcrafted feature-based support vector machine (SVM) systems. 
Andreini et al. \cite{andreini2018deep} use k-means clustering to perform foreground-background segmentation, rather than classification or counting colonies.
Multiple methods \cite{beznik2022deep,falk2019u,ramesh2018cell} approach colony detection by using modified U-Net \cite{ronneberger2015u} structures. 
Mask-RCNN \cite{he2017mask} has also been adapted multiple times \cite{liu2022two,naets2021mask} for detecting and segmenting microorganisms in agar dishes.
Majchrowska et al. \cite{majchrowska2021agar} used an image-patch approach, dividing high-resolution images into smaller overlapping areas to perform individual object detection \cite{cai2018cascade,Ren_2017} and then merging the resulting bounding boxes.
However, a common drawback of these methods is that they were developed either for low-resolution images or for image slices.

\subsection{Object detection}
In recent years, deep learning approaches have made significant progress in the field of object detection \cite{lin2017focal,Ren_2017,ebert2022multitask}, outperforming classical methods by a large margin, highlighting the potential of the current SoTA to improve accuracy and speed of colony detection.
Two stage detectors such as Faster-RCNN \cite{Ren_2017} and its variants \cite{cai2018cascade,he2017mask} first define regions of interest and then perform object detection.
RetinaNet \cite{lin2017focal} introduced Focal loss to address the class imbalance problem in one-stage detectors.
FCOS \cite{tian2019fcos} and VariFocalNet \cite{zhang2021varifocalnet} locate objects of interest by using anchor points and point-to-boundary distances. 
TOOD \cite{feng2021tood} presented a task-aligned learning strategy for explicitly aligning the two tasks of classification and localization in a learning-based manner.
All these methods have in common that they focus on the prediction-head. 
As a neck, a Feature Pyramid Network (FPN) \cite{lin2017feature} is usually used to improve accuracy by creating multi-scale features.
The Path Aggregation Feature Pyramid Network (PAFPN) \cite{liu2018path} extends the FPN approach by adding a bottom-up path to enhance FPN features with accurate localization signals from low levels.
YOLOv4 \cite{bochkovskiy2020yolov4} introduces further bottlenecks into the PAFPN for more diverse representations.
ResFPN \cite{rishav2021resfpn} enhances FPN by integrating multiple residual skip connections to leverage information from higher scales for stronger and more localized features.
The transformer-based DETR \cite{detr,zhu2020deformable} works entirely without FPN and achieves still SoTA-results.
The methods mentioned are designed for the COCO dataset \cite{lin2014microsoft}, which is known for its diversity and mainly consists of medium-sized images and objects.
Therefore, the benchmark does not adequately represent the challenge of high-resolution hygiene monitoring, with its numerous tiny colony growths and homogeneous backgrounds.
Accordingly, the aforementioned methods are only conditionally suitable for solving the task of colony detection.

To address these drawbacks, we investigate cutting-edge object detection techniques and incorporate a specialized Attention-based Path Aggregation Feature Pyramid Network (AttnPAFPN) for high-resolution feature extraction in order to detect colonies on agar dishes (see Figure \ref{fig:architecture}). 
The goal of our work is to provide a solution specific to the challenges of colony detection, improving both the accuracy and efficiency compared to SoTA methods.

\section{Method}
This section outlines the design choices of our proposed AttnPAFPN to specifically address the limitations of current SoTA methods in processing high-resolution images.
The proposed detection network consists of three key components: a backbone for extracting image features from the input, our neck (AttnPAFPN) for generating a hierarchical feature representation at different scales, followed by a detection head for the final predictions (e.g. TOOD \cite{feng2021tood}).

\subsection{AttnPAFPN}
Our primary contribution is the novel AttnPAFPN network neck, tailored to high-resolution images and small objects. AttnPAFPN utilizes our efficient-global self-attention mechanism and a new high-resolution output, allowing the network to focus on essential features, even for extremely small objects. Our streamlined method is further optimized using concepts from CSP-Net \cite{wang2020cspnet}, resulting in improved performance, lower parameter counter, and reduced complexity.
The end-to-end trainable encoder-decoder is shown in Figure \ref{fig:architecture}. 

At the initial stage of our AttnPAFPN, we use the lowest resolution backbone features (e.g. with a total stride of $32$). 
These features are passed through a CSP-Bottleneck block to create high-level features, which are then used in both the top-down and bottom-up pathways.
In the top-down pathway, the features are first upsampled by a factor of $2$, then concatenated with the backbone features of corresponding size, before being processed again by a subsequent CSP-Bottleneck.
This process is repeated until the last stage (stride of $4$) is reached, which enables AttnPAFPN to recognize tiny objects due to its high-resolution features.
To reduce computational complexity and the number of parameters, we compress the depth of the backbone features by applying a  $1\times1$ convolutional layer before passing them to the feature pyramid.
The bottom-up path of our AttnPAFPN also utilizes CSP-Bottleneck blocks, but instead of upsampling, a strided convolutional layer is used to process the features. 
This path also includes a final strided $3\times3$ convolutional layer to generate an output with a factor of $\frac{1}{64}$ of the original image size and enables the network to  recognize large objects.
Our final AttnPAFPN predicts objects at five different scales, with total strides of $\{4, 8, 16, 32, 64\}$. 

\begin{figure*}[t]
     \centering
     \begin{subfigure}[b]{0.4\textwidth}
         \centering
         \includegraphics[height=5cm]{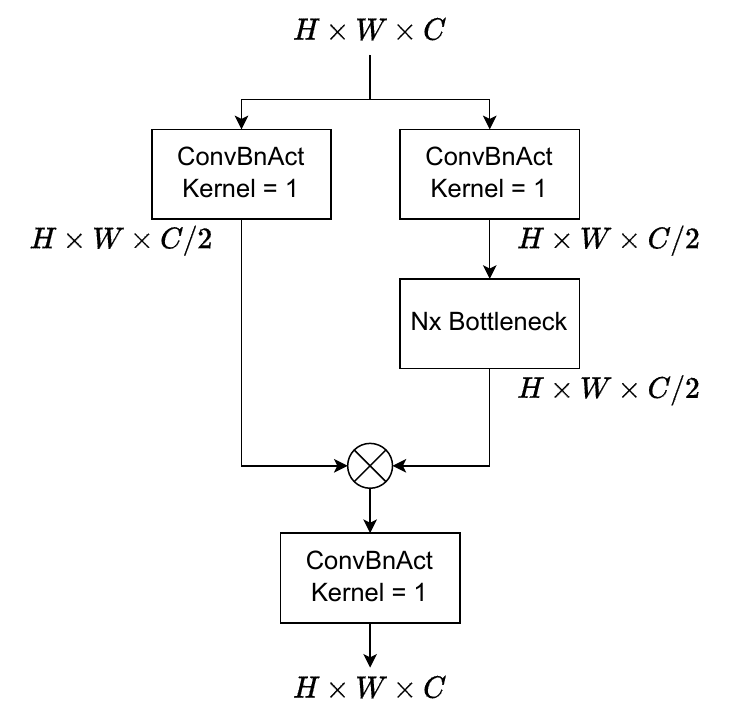}
         \caption{CSP-Bottleneck}
         \label{fig:csp_bttlneck}
     \end{subfigure}
     \hfill
     \begin{subfigure}[b]{0.25\textwidth}
         \centering
         \includegraphics[height=5cm]{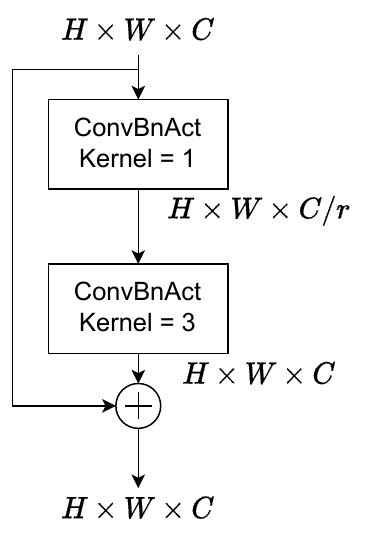}
         \caption{Bottleneck}
         \label{fig:bttlneck}
     \end{subfigure}
     \hfill
     \begin{subfigure}[b]{0.3\textwidth}
         \centering
         \includegraphics[height=5cm]{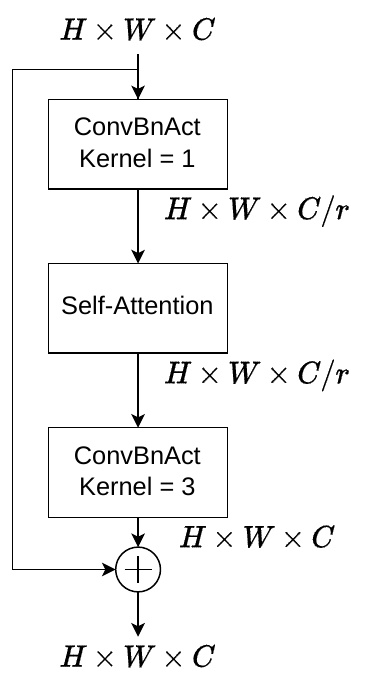}
         \caption{SA-Bottleneck}
         \label{fig:t-bttlneck}
     \end{subfigure}
        \caption{\textbf{Illustrations of the used network modules.}}
        \label{fig:modules}
\end{figure*}
\label{sec:method}

\subsection{Self-Attention augmented CSP-Bottlenecks}
One of the key contributions of our work is the integration of transformers \cite{dosovitskiy2020image,vaswani2017attention} into CSP-Bottlenecks \cite{wang2020cspnet} (Figure \ref{fig:csp_bttlneck},\ref{fig:t-bttlneck}), similar to the approach taken by BoTNet \cite{srinivas2021bottleneck} integrating transformers into ResNet for image classification \cite{he2016deep}. 
The structure of CSP-Bottlenecks can be seen in Figure \ref{fig:csp_bttlneck}. First, the incoming featuremaps are divided into two parts in depth. The first part is passed directly to the output after a single pointwise-convolution operation. The other half is processed $N$ times by a residual bottleneck (see Figure \ref{fig:bttlneck}) and then concatenated with the first half. Finally, a pointwise convolution is performed to enable communication between the channels.
To integrate self-attention mechanisms into these structure, we replace the convolutional bottleneck with our self-attention augmented version (see Figure \ref{fig:t-bttlneck}.
However, the use of standard self-attention is limited by its quadratic complexity, especially when applied to high-resolution images, such as those of hygiene monitoring.
To address this challenge, we compare two resolution-optimized transformers: our novel efficient-global self-attention and local-window self-attention similar to Swin Transformer \cite{liu2021swin}.
In general, a transformer-layer \cite{vaswani2017attention} can be described as 
\begin{equation}\label{eq:transformer}
\textstyle
\begin{split} 
y^{*} &=\text{Self-Attention}(\text{LN(} x)) + x, \\
y &=\text{FFN}(\text{LN(} y^{*})) + y^{*} ,
\end{split} 
\end{equation}
with $x$ as its input and $y$ as output features. LN refers to layer normalization \cite{ba2016layer}, and FFN to a linear feed-forward layer. Self-attention \cite{liu2021swin} can be formulated as
\begin{equation}\label{eq:mhsa}
\textstyle
\text{Self-Attention}(q,k,v) = \text{Softmax}(\frac{qk^T}{\sqrt{d}}+b)v,
\end{equation}
where $q, k, v$ are query, key and value matrices generated from input-features, $d$ is a scaling factor and $b$ is a trainable relative position bias term.
Inspired by SegFormer \cite{xie2021segformer}, we extend our feed-forward-network (FFN) by CNN-layers, adding an inductive bias for finer localization using additional positional information: 
\begin{equation}\label{eq:ccf-ffn}
\textstyle
\begin{split}
    y^{*} &= \text{GeLU}(\text{LN}(\text{PWConv}(x))), \\
    y &= \text{PWConv}(\text{GeLU}(\text{LN}(\text{DWConv}_{3\times3}(y^{*})))) + x, \\
    \end{split}
\end{equation}
where GeLU \cite{hendrycks2016gaussian} corresponds to Gaussian Error Linear Unit activation, PWConv to a point-wise convolution and $\text{DWConv}_{3\times3}$ to a depth-wise $3\times3$ convolution.
 
Local-window self-attention splits input features into non-overlapping windows with limited receptive fields, before applying multihead self-attention.
As a result, the computational effort of self-attention is linear to the window-size. 
One downside is that information cannot pass between the windows within a layer.
Several successive layers with shifting windows is necessary to create a global receptive field.
In our experiments we follow the window partitioning strategy of Swin Transformer  \cite{liu2021swin}.
In contrast, our efficient-global self-attention reduces the spatial resolution of the input to a fixed global size by performing adaptive max-pooling on the input. 
The size of the global window is freely selectable, but we have set the window size in all our networks to $\frac{1}{64}$ of the original resolution. 
In case of $1024 \times 1024$ resolution, the fixed global window would be $16 \times 16$.
Regardless of the input resolution, it is also possible to set the global window to a fixed size. 
This results in a network complexity that is completely independent of the image resolution.
With our efficient-global self-attention we create a single window with a global receptive field to which self-attention is subsequently applied.
These and many more transformer variants can be easily inserted into the bottleneck structure as shown in Figure \ref{fig:t-bttlneck}. 

\section{Evaluation}
\label{sec:eval}

\begin{table*}[t]
\scriptsize
\centering 
\caption{\textbf{Ablation study} on the effectiveness of the components of our AttnPAFPN. TOOD \cite{feng2021tood} is used as head for evaluation on AGAR-dataset \cite{majchrowska2021agar}. We evaluated onthe  high-resolution and low-resolution subset with a image-size of $1536\times1536$.}
\label{tab:ablation}
\begin{tabular}{l|c|cccc|cccc}
\hline
\multicolumn{1}{c|}{\multirow{2}{*}{\textbf{Method}}} & \multirow{2}{*}{\textbf{Params}} & \multicolumn{4}{c|}{\textbf{HR-Subset}}                                                & \multicolumn{4}{c}{\textbf{LR-Subset}}                                                 \\
\multicolumn{1}{c|}{}                                 &                                 & \multicolumn{1}{c}{\textbf{mAP}} & \multicolumn{1}{c}{\textbf{AP\textsuperscript{50}}} & \multicolumn{1}{c}{\textbf{AP\textsuperscript{75}}} & \textbf{ R\textsuperscript{50} } & \multicolumn{1}{c}{\textbf{mAP}} & \multicolumn{1}{c}{\textbf{AP\textsuperscript{50}}} & \multicolumn{1}{c}{\textbf{AP\textsuperscript{75}}} & \textbf{ R\textsuperscript{50} } \\ \hline
TOOD-Baseline \cite{feng2021tood}                                        &  \textbf{32.0 M}                               & 57.7                              & 82.6                               & 68.2      & 83.0     & 67.2                                  & 95.9                                    & 80.0      & 96.5              \\ 
+ CSP-PAFPN                                               & 74.3 M                                & 63.3                              & 90.2                               & 74.8      & 90.8    & 68.0                                   &  95.8                                  & 81.1       & 96.5       \\
+ Attn-Bottleneck                                     & 64.4 M                                & 66.5                              & 95.3                               & 78.1      & 96.0    &  69.1                                 & 97.5                                   & 82.6          & 98.3     \\
+ Extra Detection-Scales                              & 66.3 M                                & 67.7                              & 96.1                               & 80.3     &   \textbf{96.8}   & \textbf{69.7}                                  & \textbf{98.0}                                    & \textbf{83.8}         &    \textbf{98.9} \\
+ Feature-compression-layer                              & 32.8 M  & 68.1 & 96.2 & 81.0 & \textbf{96.8} & 69.3  & 97.6  & 82.8 &  98.4             \\
+ Multiscale-Training                                 & 32.8 M                                & \textbf{68.2}                                  & \textbf{96.3}                                   & \textbf{81.1}   & \textbf{96.8}            & 69.5                                  & \textbf{98.0}                                   & 83.4   & 98.6                      \\ \hline
\end{tabular}
\end{table*}

\begin{table*}[t]
\scriptsize
\centering
\caption{\textbf{Ablation study} on the effectiveness of the local-window (v1) and efficient-global self-attention (v2) for our AttnPAFPN. TOOD \cite{feng2021tood} is used as the detection head for evaluation on the AGAR-dataset \cite{majchrowska2021agar}. We evaluated on both subsets with a image-size of $1536\times1536$.}
\label{tab:ablationTrans}
\begin{tabular}{l|c|cccc|cccc}
\hline
\multicolumn{1}{c|}{\multirow{2}{*}{\textbf{Method}}} & \multirow{2}{*}{\textbf{Params}} & \multicolumn{4}{c|}{\textbf{HR-Subset}}                                                & \multicolumn{4}{c}{\textbf{LR-Subset}}                                                 \\
\multicolumn{1}{c|}{}                                 &                                 & \multicolumn{1}{c}{\textbf{mAP}} & \multicolumn{1}{c}{\textbf{AP\textsuperscript{50}}} & \multicolumn{1}{c}{\textbf{AP\textsuperscript{75}}} & \textbf{ R\textsuperscript{50} } & \multicolumn{1}{c}{\textbf{mAP}} & \multicolumn{1}{c}{\textbf{AP\textsuperscript{50}}} & \multicolumn{1}{c}{\textbf{AP\textsuperscript{75}}}& \textbf{ R\textsuperscript{50} } \\ \hline
TOOD-Baseline \cite{feng2021tood}                                         & \textbf{32.0 M}                                & 57.7                              & 82.6                               & 68.2       & 83.0  & 67.2                                  & 95.9                                    & 80.0           &   96.5 \\ \hline
+ AttnPAFPNv1                                         & 84.1 M                                & 66.5                              & 95.1                               & 78.5    &  95.7    & 69.0                                   & 97.4                                   & 82.6       &   98.1    \\
+ Extra Detection-Scales                                          & 87.0 M                                & 67.2                              & 95.7                               & 79.5    &   96.5   & \textbf{69.7}                                  & \textbf{98.0}                                   & 83.3       &  98.8     \\ \hline
+ AttnPAFPNv2                                         & 64.4 M                                 & 66.5                              & 95.3                               & 78.1     &   96.0  &  69.1                                 & 97.5                                   & 82.6         & 98.3    \\
+ Extra Detection-Scales                                          & 66.3 M                                 & \textbf{67.7}                              & \textbf{96.1}                               & \textbf{80.3}    &   \textbf{96.8}   & \textbf{69.7}                                  & \textbf{98.0}                                 & \textbf{83.8}          &  \textbf{98.9}  \\ \hline
\end{tabular}
\end{table*}

Our evaluation focuses on demonstrating the benefits of our AttnPAFPN in high-resolution object detection. For this purpose we use the public AGAR dataset \cite{majchrowska2021agar}, containing high-resolution images of five different types of bacteria on agar plates. The data is divided into the higher-resolution (HR) and lower-resolution subsets (LR). 
The HR subset contains approximately 5k training images and 2k test images, with a resolution of around $4,000^2$ pixels. The LR subset has around 3.5k training images and 1k test images with a resolution of $2,048^2$ pixels. 
In addition to these two subsets, a third mixed-resolution subset is created by combining both subsets.

We perform an ablation study to determine the effect individual components have on our methods accuracy. Results are shown in Tables \ref{tab:ablation} and \ref{tab:ablationTrans}. 
%In order to show the general applicability and benefits of our method, we implement AttnPAFPN in current state-of-the-art methods (e.g. TOOD \cite{feng2021tood}). Results are listed in Table \ref{tab:eval}.
Furthermore, we implement our AttnPAFPN in current SoTA methods (e.g. TOOD \cite{feng2021tood}) and compare it with five different object detection models, listing the results in Table \ref{tab:eval}.  
%We implement all methods in the MMDetection-Framework \cite{chen2019mmdetection}.
%We use the mAP metric to evaluate the performance of all our models. 
All methods are implemented in the MMDetection-Framework \cite{chen2019mmdetection} and we use the mAP metric to evaluate their performance.
mAP provides a comprehensive assessment of accuracy and recall, averaging the maximum precision score for each recall value of all classes.
In hygiene monitoring, detecting all colonies is a priority over precise localization. 
Hence, we use the Recall at an IoU threshold of 0.5 (R\textsuperscript{50}) as an additional metric.

\subsection{Ablation Study}
\label{sec:ablation}

In our first experiment, we assess the impact of our method by comparing AttnPAFPN with a baseline model (TOOD \cite{feng2021tood} + FPN \cite{lin2017feature}) as shown in Table \ref{tab:ablation}. 
All networks are trained for 20 epochs with the SGD optimizer, a batch-size of 8, and use a pre-trained ResNet50 \cite{he2016deep} as their backbone. 
The learning rate starts at $5 \cdot 10^{-3}$ and decreases by a factor of 10 after 8 and 16 epochs.

Replacing the standard FPN in TOOD with the convolutional CSP-PAFPN leads to an improvement in mAP ($+5.6 / +0.8$) and Recall ($+7.8 / \pm 0$), but also increases the number of parameters by more than 100\%.
By introducing our efficient-global self attention (SA) into the CSP-Bottlenecks, we were able to reduce the parameters by over 15\% and further boost mAP ($+3.3/+1.1$) and R\textsuperscript{50} ($+5.2/+1.8$) compared to the previous step. 
In these initial experiments, all network necks use only the backbone scales $\{8,16,32\}$ for predictions.
To ensure better recognition of particularly large and tiny colonies, we add two more scales, so that we ultimately perform detection across five resolutions: $\{4,8,16,32,64\}$.
To address the heavy-weight nature of our network, we implemented $1\times 1$ feature-compression layers in our AttnPAFPN, reducing the depth $C$ of backbone-features to $C^{*}=256$. 
Through this feature reduction our method achieves a parameter count comparable to the baseline FPN, while still achieving a stronger performance in terms of mAP and R\textsuperscript{50}. 
For a final increase in performance, we utilize multi-scale training.
Overall AttnPAFPN increases mAP by $+10.5 / +2.3$ and Recall by $+13.8 / +2.1$ in comparison to the baseline.

\begin{table*}[t]
\scriptsize
\centering
\caption{\textbf{Comparison of detection accuracy} on the AGAR \cite{majchrowska2021agar} validation set. Different SoTA methods \cite{feng2021tood,lin2017focal,Ren_2017,zhang2021varifocalnet,zhu2020deformable} are compared with our approach. We use TOOD \cite{feng2021tood} and Faster-RCNN \cite{Ren_2017} as head. The comparison is performed with different data-subsets.}
\label{tab:eval}
\begin{tabular}{lccccccccccccc}
\hline
\multicolumn{1}{c|}{}                                  & \multicolumn{1}{c|}{}                                 & \multicolumn{4}{c|}{\textbf{HR-Subset}}                                                                                & \multicolumn{4}{c|}{\textbf{LR-Subset}}                                                                                & \multicolumn{4}{c}{\textbf{MR-Subset}}                                         \\
\multicolumn{1}{c|}{\multirow{-2}{*}{\textbf{Method}}} & \multicolumn{1}{c|}{\multirow{-2}{*}{\textbf{Params}}} & \multicolumn{1}{c}{\textbf{mAP}} & \multicolumn{1}{c}{\textbf{AP\textsuperscript{50}}} & \multicolumn{1}{c}{\textbf{AP\textsuperscript{75}}} & \multicolumn{1}{c|}{\textbf{ R\textsuperscript{50} }}            & \multicolumn{1}{c}{\textbf{mAP}} & \multicolumn{1}{c}{\textbf{AP\textsuperscript{50}}} & \multicolumn{1}{c}{\textbf{AP\textsuperscript{75}}} & \multicolumn{1}{c|}{\textbf{ R\textsuperscript{50} }}            & \multicolumn{1}{c}{\textbf{mAP}} & \multicolumn{1}{c}{\textbf{AP\textsuperscript{50}}} & \multicolumn{1}{c}{\textbf{AP\textsuperscript{75}}} & \multicolumn{1}{c}{\textbf{ R\textsuperscript{50} }}            \\ \hline
\multicolumn{14}{c}{Patches: $512 \times 512$}\\ \hline
\multicolumn{1}{l|}{Faster-RCNN \cite{Ren_2017,majchrowska2021agar}}                       & \multicolumn{1}{c|}{41.5 M}                                 & 49.3                                  & 76.7                & 54.8                    & \multicolumn{1}{c|}{-}                         & 56.0                                   & 86.5      & 63.6                             & \multicolumn{1}{c|}{-}                         & -                         & -                         & -           & -               \\
\multicolumn{1}{l|}{Cascade-RCNN \cite{cai2018cascade,majchrowska2021agar}}                         & \multicolumn{1}{c|}{69.2 M}                                 &  51.6                                 & 79.2                  & 57.0                 & \multicolumn{1}{c|}{-}                         & 58.4                                   & 88.6  & 68.3                                 & \multicolumn{1}{c|}{-}                         & -                         & -                         & -               & -          \\ \hline
\multicolumn{14}{c}{$1024 \times 1024$}\\ \hline
\multicolumn{1}{l|}{Faster-RCNN \cite{Ren_2017}}                       & \multicolumn{1}{c|}{41.5 M}                                 & 45.7                                  & 65.6             & 53.9                       & \multicolumn{1}{c|}{65.9}                         & 62.2                                   & 89.7            & 74.3                       & \multicolumn{1}{c|}{90.1}                         & 50.0                         & 71.7              & 59.4           & 72.0                          \\
\multicolumn{1}{l|}{RetinaNet \cite{lin2017focal}}                         & \multicolumn{1}{c|}{37.7 M}                                 &  42.5                                 & 68.1                     & 46.9              & \multicolumn{1}{c|}{74.5}                         & 59.2                                   & 90.8        & 67.9                           & \multicolumn{1}{c|}{93.7}                         & 50.6                         & 77.1           & 58.2              & 81.9                         \\
\multicolumn{1}{l|}{TOOD \cite{feng2021tood}}                              & \multicolumn{1}{c|}{32.0 M}                                 & 57.3                                   & 82.4                & 67.6                   & \multicolumn{1}{c|}{82.5}                         & 66.9                                  & 95.4                 & 79.1                  & \multicolumn{1}{c|}{96.0}                         & 59.8                         &  85.8           & 70.5             & 86.3                         \\
\multicolumn{1}{l|}{Def. DETR \cite{zhu2020deformable}}                              & \multicolumn{1}{c|}{41.3 M}                                 & 49.8                                  & 81.3                    & 55.0               & \multicolumn{1}{c|}{82.8}                         & 64.4                                  & 95.2          & 76.4                         & \multicolumn{1}{c|}{96.5}                         & 57.3                         & 86.3                 & 66.8        & 87.2                          \\
\multicolumn{1}{l|}{VariFocalNet \cite{zhang2021varifocalnet}}                      & \multicolumn{1}{c|}{32.7 M}                                 & 56.4              & 81.4    & 66.0           & \multicolumn{1}{c|}{82.1}                        & 66.5              & 94.8       & 80.2        & \multicolumn{1}{c|}{95.6}                         & 59.5     & 85.1 & 73.6     & 85.8     \\

\rowcolor[HTML]{EFEFEF} 
\multicolumn{1}{l|}{Faster-RCNN \cite{Ren_2017} + Ours}      & \multicolumn{1}{c|}{42.7 M}         & 49.1                                  & 72.4     & 57.7                              & \multicolumn{1}{c|}{73.0} & 62.6                                  & 91.3         &              75.0              & \multicolumn{1}{c|}{91.7} & 52.5  & 77.2 & 61.9 & 77.6 \\ 
\rowcolor[HTML]{EFEFEF} 
\multicolumn{1}{l|}{TOOD \cite{feng2021tood} + Ours}      & \multicolumn{1}{c|}{32.8 M}         & 67.5                                  & 95.8      & 80.6                             & \multicolumn{1}{c|}{95.8} & 68.9                                  &  97.6          & 82.6                        & \multicolumn{1}{c|}{98.4} & 68.4  & 96.6 & 81.4 & 96.6 \\ 

\hline
\multicolumn{14}{c}{$1536 \times 1536$}\\ \hline
\multicolumn{1}{l|}{Faster-RCNN \cite{Ren_2017}}                       & \multicolumn{1}{c|}{41.5 M}                                 & 56.0                                   & 80.2           & 66.0                        & \multicolumn{1}{c|}{80.8}                         & 64.7                                  & 93.3   & 77.0                                & \multicolumn{1}{c|}{93.8}                         & 57.9                         & 82.9            & 68.4             & 83.3                         \\
\multicolumn{1}{l|}{RetinaNet \cite{lin2017focal}}                         & \multicolumn{1}{c|}{37.7 M}                                 & 50.3                                  & 77.7         & 57.1                          & \multicolumn{1}{c|}{80.2}                         & 59.5                                  & 90.8        & 68.8                           & \multicolumn{1}{c|}{92.7}                         &  54.6                        & 82.6           & 62.9              & 84.2                         \\
\multicolumn{1}{l|}{TOOD \cite{feng2021tood}}                              & \multicolumn{1}{c|}{32.0 M}                                 & 57.7                                  & 82.6            & 68.2                       &  \multicolumn{1}{c|}{83.0}                         & 67.2                                  & 95.9              & 80.0                & \multicolumn{1}{c|}{96.5}                         & 61.8                        & 86.8             & 73.9            & 87.3                        \\
\multicolumn{1}{l|}{Def. DETR \cite{zhu2020deformable}}                              & \multicolumn{1}{c|}{41.3 M}                                 & 51.9                                   &  82.2                  & 58.4                &  \multicolumn{1}{c|}{83.5}                         & 65.3                                  & 94.9             & 76.8                      & \multicolumn{1}{c|}{96.6}                         & 56.8                          & 86.5            & 66.1             &  87.3                        \\
\multicolumn{1}{l|}{VariFocalNet \cite{zhang2021varifocalnet}}                      & \multicolumn{1}{c|}{32.7 M}                                 & \multicolumn{1}{c}{59.7}              & \multicolumn{1}{c}{83.1}      & 71.0         & \multicolumn{1}{c|}{83.6}                         & \multicolumn{1}{c}{67.6}              & \multicolumn{1}{c}{95.8}    & 80.2           & \multicolumn{1}{c|}{96.6}                         & \multicolumn{1}{c}{61.8}     & \multicolumn{1}{c}{86.4}  & 73.6   & \multicolumn{1}{c}{87.0}     \\
\rowcolor[HTML]{EFEFEF} \multicolumn{1}{l|}{Faster-RCNN \cite{Ren_2017} + Ours}      & \multicolumn{1}{c|}{42.7 M}         & 61.5                                   & 89.2                 & 72.4                & \multicolumn{1}{c|}{89.8} & 65.8                                  & 95.1                            & 79.0                                       & \multicolumn{1}{c|}{95.1} & 62.6  &  91.2 & 74.3 & 91.7 \\ 

\rowcolor[HTML]{EFEFEF} \multicolumn{1}{l|}{TOOD \cite{feng2021tood} + Ours}      & \multicolumn{1}{c|}{32.8 M}         & 68.2                                   & 96.3                         & 81.1         & \multicolumn{1}{c|}{96.2} & 69.5                                  & 98.0             & 83.4                                                      & \multicolumn{1}{c|}{98.7} & 68.0  & 96.2  & 81.6 & 97.0 \\

\hline

\multicolumn{14}{c}{$2048 \times 2048$} \\ \hline
\multicolumn{1}{l|}{Faster-RCNN \cite{Ren_2017}}                       & \multicolumn{1}{c|}{41.5 M}                                 & 58.0                                   & 82.2          & 68.5                          & \multicolumn{1}{c|}{82.5}                         & 66.6                                  & 95.4              & 79.4                     & \multicolumn{1}{c|}{95.9}                         & 60.2                         & 85.6             & 71.6            & 85.9                         \\
\multicolumn{1}{l|}{RetinaNet \cite{lin2017focal}}                         & \multicolumn{1}{c|}{37.7 M}                                 & 56.0                                  & 81.9                    & 65.2              & \multicolumn{1}{c|}{83.1}                         & 64.2                                  & 93.8          & 75.6                         & \multicolumn{1}{c|}{95.0}                         & 56.4                         & 84.1         & 65.3               &  85.4                       \\
\multicolumn{1}{l|}{TOOD \cite{feng2021tood}}                              & \multicolumn{1}{c|}{32.0 M}                                 & 60.6                                  & 84.3                  & 72.5                 & \multicolumn{1}{c|}{84.6}                         &  67.4                                 &  95.9            & 80.5                      & \multicolumn{1}{c|}{96.5}                         & 62.7                         & 87.4          & 75.2               & 87.8                         \\
\multicolumn{1}{l|}{Def. DETR \cite{zhu2020deformable}}                              & \multicolumn{1}{c|}{41.3 M}                                 &  53.9                                 &  83.0            & 53.9                      & \multicolumn{1}{c|}{83.9}                         & 64.9                                  & 95.5                 & 76.4                  & \multicolumn{1}{c|}{96.8}                         &  57.8                        & 86.8            & 67.5             & 87.4                          \\
\multicolumn{1}{l|}{VariFocalNet \cite{zhang2021varifocalnet}}                      & \multicolumn{1}{c|}{32.7 M}                                 & \multicolumn{1}{c}{60.5}              & \multicolumn{1}{c}{83.6}     & 72.2          & \multicolumn{1}{c|}{84.1}                         & \multicolumn{1}{c}{68.2}              & \multicolumn{1}{c}{95.9}    & 81.0           & \multicolumn{1}{c|}{96.7}                         & \multicolumn{1}{c}{63.0}     & \multicolumn{1}{c}{87.0} & 74.9    & \multicolumn{1}{c}{87.5}     \\
\rowcolor[HTML]{EFEFEF} 
\multicolumn{1}{l|}{Faster-RCNN \cite{Ren_2017} + Ours}      & \multicolumn{1}{c|}{42.7 M}         & 64.0                                  & 93.0            & 75.3                       & \multicolumn{1}{c|}{93.5} & 67.5                                  & 96.9          & 81.2                          & \multicolumn{1}{c|}{97.3} & 64.4  & 93.7 & 76.4 & 94.1  \\ 
\rowcolor[HTML]{EFEFEF} 
\multicolumn{1}{l|}{TOOD \cite{feng2021tood} + Ours}      & \multicolumn{1}{c|}{32.8 M}         & 68.9                                  & 96.8       & 82.1                            & \multicolumn{1}{c|}{97.5} &  70.5                                 & 98.1              & 84.7                     & \multicolumn{1}{c|}{98.9} & 68.4   & 96.1 & 82.0 & 97.4  \\ 
\hline
\end{tabular}
\end{table*}

In Section \ref{sec:method} of our study, we present two variants of efficient transformer layers that are specifically designed for high-resolution images. 
Table \ref{tab:ablationTrans} compares the performance of local-window SA (v1) and efficient-global SA (v2). 
The results indicate that efficient-global SA, which provides a coarse-grained overview of the entire image, leads to a significant improvement in accuracy. 
The differences in mAP are only marginal on the HR subset; on the LR subset, both networks achieve almost identical accuracy. 
The decisive point here is the significantly lower complexity and the lower number of weights of the global self-attention.

\subsection{Quantitative Evaluation} \label{sec:quant}
In our final experiment, as listed in Table \ref{tab:eval}, we compare the performance of our proposed method, AttnPAFPN, with SoTA object detection methods \cite{feng2021tood,lin2017focal,Ren_2017,zhang2021varifocalnet,zhu2020deformable}.
The training process of all the networks is equal to the description in Section \ref{sec:ablation}.
The first few rows which are titled with "Patches: $512 \times 512$" present the results of Majchrowska et al. \cite{majchrowska2021agar}. 
They divide the images into patches of size $512 \times 512$ and then detect the colonies in each of these patches using Faster-RCNN \cite{Ren_2017} and Cascade-RCNN \cite{cai2018cascade} with ResNet50 \cite{he2016deep} as the backbone, similar to our setup. 
The following lines contain the results of the SoTA and our method using the full image under different resolutions.
Upon comparison with Faster-RCNN, our AttnPAFPN shows lower performance for lower resolution, especially $1024\times1024$ for the HR-Subset. 
However, as the resolution increases, AttnPAFPN outperforms all baselines by a large margin. 
Furthermore, our AttnPAFPN achieved best results for TOOD \cite{feng2021tood} at a final resolution of $2048\times2048$, but it also shows excellent results even at moderate resolutions and therefore does not necessarily require very high resolutions with high computational overhead.

\subsection{Further Experiments}
Extending the evaluations in Section \ref{sec:ablation} and \ref{sec:quant}, we perform several more experiments on the AGAR dataset \cite{majchrowska2021agar}.
We investigating ability of generalization only using a small number of training data and examined various backbones.
Furthermore, we evaluated the performace of our network on COCO \cite{lin2014microsoft} for general object detection and on LIVECell \cite{edlund2021livecell} for detection of cells on low-resolution images.

\subsubsection{Limited Data Analysis}

\begin{table}[t]
\scriptsize
\centering
\caption{\textbf{Comparison of detection accuracy} of our method to the SoTA on the AGAR \cite{majchrowska2021agar} validation. For training, different data-splits with 10 \%, 5\% and 1\% of the original 5000 training images are used. For this experiment, we use TOOD \cite{feng2021tood} and Faster-RCNN \cite{Ren_2017} as the network heads and evaluate at a resolution of $1536 \times 1536$.}
\label{tab:data}
\begin{tabular}{lccccc}
\hline
\multicolumn{1}{c|}{\multirow{2}{*}{\textbf{Method}}} & \multicolumn{1}{c|}{\multirow{2}{*}{\textbf{Params}}} & \multicolumn{4}{c}{\textbf{Metrics}}                                                                                                        \\
\multicolumn{1}{c|}{}                                 & \multicolumn{1}{c|}{}                                 & \multicolumn{1}{c}{\textbf{mAP}} & \multicolumn{1}{c}{\textbf{AP\textsuperscript{50}}} & \multicolumn{1}{c}{\textbf{AP\textsuperscript{75}}} & \multicolumn{1}{c}{\textbf{R\textsuperscript{50}}} \\ \hline
\multicolumn{6}{c}{\textbf{Subset 10 \%}}                                                                                                                                                                                                                  \\ \hline
\multicolumn{1}{l|}{Faster-RCNN \cite{Ren_2017}}                      & \multicolumn{1}{c|}{41.5 M}                           &    53.4                            &     78.8                              &        62.3                           &  79.4                                \\
\multicolumn{1}{l|}{TOOD \cite{feng2021tood}}                             & \multicolumn{1}{c|}{32.0 M}                           & 54.0                             & 80.1                              & 63.2                              & 81.9                             \\
\multicolumn{1}{l|}{Faster-RCNN \cite{Ren_2017} + ours}               & \multicolumn{1}{c|}{42.7 M}                         &     57.0        &  87.2            & 65.7          & 88.2             \\
\multicolumn{1}{l|}{TOOD \cite{feng2021tood} + ours}                      & \multicolumn{1}{c|}{32.8 M}                           & 62.9                             & 92.3                              & 73.8                              & 93.8                             \\ \hline
\multicolumn{6}{c}{\textbf{Subset 5 \%}}                                                                                                                                                                                                                   \\ \hline
\multicolumn{1}{l|}{Faster-RCNN \cite{Ren_2017}}                      & \multicolumn{1}{c|}{41.5 M}                           &  51.7                                &        77.7                           &       59.8                            &     78.6                             \\
\multicolumn{1}{l|}{TOOD \cite{feng2021tood}}                             & \multicolumn{1}{c|}{32.0 M}                           & 51.3                             & 77.6                              & 59.4                              & 79.5                             \\
\multicolumn{1}{l|}{Faster-RCNN \cite{Ren_2017} + ours}               & \multicolumn{1}{c|}{42.7 M}                           & 55.4             & 86.4              & 63.6             & 87.9             \\
\multicolumn{1}{l|}{TOOD \cite{feng2021tood} + ours}                      & \multicolumn{1}{c|}{32.8 M}                           & 61.8                             & 92.0                              & 72.5                              & 94.1                             \\ \hline
\multicolumn{6}{c}{\textbf{Subset 1 \%}}                                                                                                                                                                                                                   \\ \hline
\multicolumn{1}{l|}{Faster-RCNN \cite{Ren_2017}}                      & \multicolumn{1}{c|}{41.5 M}                           & 41.2                              &     70.6                              &        43.6                           &  73.7                   \\
\multicolumn{1}{l|}{TOOD \cite{feng2021tood}}                             & \multicolumn{1}{c|}{32.0 M}                           & 36.8                             & 61.2                              & 40.5                              & 68.3                             \\
\multicolumn{1}{l|}{Faster-RCNN \cite{Ren_2017} + ours}               & \multicolumn{1}{c|}{42.7 M}                           &   42.7         &  75.4            & 43.7          & 84.9                            \\
\multicolumn{1}{l|}{TOOD \cite{feng2021tood} + ours}                      & \multicolumn{1}{c|}{32.8 M}                           & 42.6                             & 72.5                              & 46.1                              & 80.9                             \\ \hline
\end{tabular}
\vspace{-3mm}
\end{table}

In our first additional experiment, we investigate how a reduction of the amount of data affects the training of our networks.
For this reason, we created three evenly distributed subsets from the higher-resolution (HR) set, each containing 10 \% (524 images), 5 \% (262 images), and 1 \% (53 images) of the training data. 
For evaluation, we use the complete validation set of the HR subset as described in Section \ref{sec:eval}.
In contrast to the training in Section \ref{sec:quant}, we increase the number of epochs to 100 and reduce the learning rate after 50 and 80 epochs by a factor of 10.

The results listed in Table \ref{tab:data} show a drop between 3 \% to 5 \% of the mAP with respect to networks, trained on all data when using 10 \% of the training data.
The drop from TOOD \cite{feng2021tood} extended by our AttnPAFPN shows a larger loss in mAP due to the added complexity of the data-hungry transformer layers, but it still shows better accuracy than the pure TOOD trained on all data.
When using 5 \% of the training data, a similar picture emerges.
When training with only 1 \% of the image data, a very strong drop in accuracy (approximately 20 \% to 25 \%) of all networks can be seen. 
However, our AttnPAFPN still shows an above-average performance here.

\subsubsection{Backbone Analysis}

\begin{table}[t]
\scriptsize
\centering
\caption{\textbf{Comparison of detection accuracy} of our method on the AGAR \cite{majchrowska2021agar} val set. Different backbones \cite{ebert2023plg, he2016deep, liu2021swin, wang2022pvt, zhu2019deformable} are compared. For this experiment, we use TOOD \cite{feng2021tood} as head and evaluate at a resolution of $1536 \times 1536$.}
\label{tab:backbone}
\begin{tabular}{l|c|cccc}
\hline
\multicolumn{1}{c|}{\multirow{2}{*}{\textbf{Method}}} & \multirow{2}{*}{\textbf{Params}} & \multicolumn{4}{c}{\textbf{Metrics}}                        \\
\multicolumn{1}{c|}{}                                 &                                  & \textbf{mAP} & \textbf{AP\textsuperscript{50}} & \textbf{AP\textsuperscript{75}} & \textbf{R\textsuperscript{50}} \\ \hline
ResNet50 \cite{he2016deep}                                              & \textbf{32.8 M }                          & 68.2         & 96.3          & 81.1          & 96.2         \\
ResNet101-dcnv2 \cite{zhu2019deformable}                                          & 54.3 M                           &    69.2          & 96.9              &     82.9          &    97.5          \\
Swin Tiny \cite{liu2021swin}                                                & 36.4 M                           &   69.9           &   96.8            &   84.1            &   97.3           \\
PVTv2-b2 \cite{wang2022pvt}                                              & 33.7 M                           &  70.2            &    96.8           &     83.9          &     97.4         \\
PLG-ViT Tiny \cite{ebert2023plg}                                            & 34.8 M                           & \textbf{70.4}         & \textbf{97.0}          & \textbf{84.2}          & \textbf{97.6}         \\ \hline
\end{tabular}
\end{table}

During all previous experiments we have used a pretrained ResNet50 \cite{he2016deep} as the network backbone, since it is still considered as one of the most important baselines in computer vision. 
Further improvements in accuaracy can be achieved by using modern CNNs or transformer backbones.
For this reason we want to compare ResNet50 with a stronger deformable convolution backbone (ResNet101-dcnv2) \cite{zhu2019deformable} and three transformer-based backbones.
For the transformer backbones we use Swin-T \cite{liu2021swin}, PVTv2-b2 \cite{wang2022pvt}, and the high-resolution optimized PLG-ViT-T \cite{ebert2023plg, ebert2023light}.
All transformer backbones are similar in size to ResNet50 and training takes place exclusively on the higher-resolution subset of AGAR \cite{majchrowska2021agar} at a resolution of $1536 \times 1536$.
We trained ResNet101-dcnv2 with the same hyperparameters as ResNet50. 
For the transformer backbones, we adapted the training recipes proposed by the authors from COCO \cite{lin2014microsoft} to AGAR.

The results in Table \ref{tab:backbone} confirm the trend of recent years, with transformers outperforming their CNN counterparts. Even the larger ResNet101-dcnv2 backbone cannot keep up with the transformers. 
These manage to outperform ResNet50 and ResNet101-dcnv2 by about $+2$ and $+1$ mAP, respectively. 
It is also shown that the differences between transformer networks in terms of accuracy are small.  
However, this experiment shows the major drawback of the standard SA used by PVTv2. 
Even if the number of parameters is the same, the computational cost is significantly higher compared to Swin and PLG-ViT. PVTv2 requires about 200 \% more GPU memory than the other two networks during training. 
The computational effort is also significantly higher during the inference \cite{ebert2023plg}.
For this reason, PLG-ViT will be used as the backbone of choice in the final experiment to achieve the best possible trade-off between accuracy and performance.

\subsubsection{Beyond Colony Detection}

\begin{table}[t]
\scriptsize
\centering
\caption{\textbf{Comparison of detection and segmentation accuracy} on the COCO \cite{lin2014microsoft} validation set. Different methods \cite{feng2021tood,he2017mask} are compared with our approach. We use TOOD \cite{feng2021tood} and Mask-RCNN \cite{he2017mask} as the head and ResNet50 \cite{he2016deep} and PLG-ViT \cite{ebert2023plg} as the backbone.}
\label{tab:coco}
\begin{tabular}{l|c|c|cc}
\hline
\multicolumn{1}{c|}{\multirow{2}{*}{\textbf{Method}}}  & \multicolumn{1}{c|}{\multirow{2}{*}{\textbf{Backbone}}} & \multicolumn{1}{c|}{\multirow{2}{*}{\textbf{Params}}} & \multicolumn{2}{c}{\textbf{Metrics}} \\
\multicolumn{1}{c|}{}                                 & \multicolumn{1}{c|}{}   & \multicolumn{1}{c|}{}                               & \textbf{mAP\textsuperscript{bb}} & \textbf{mAP\textsuperscript{seg}}     \\ \hline
\multicolumn{1}{l|}{TOOD \cite{feng2021tood}}                 & \multicolumn{1}{l|}{ResNet50 \cite{he2016deep}}           & \multicolumn{1}{c|}{32.0 M}                           & 42.4             & -                 \\
\multicolumn{1}{l|}{TOOD \cite{feng2021tood} + ours}          & \multicolumn{1}{l|}{ResNet50 \cite{he2016deep}}            & \multicolumn{1}{c|}{32.8 M}                           &   42.6               & -                 \\
\multicolumn{1}{l|}{TOOD \cite{feng2021tood} + ours}           & \multicolumn{1}{l|}{PLG-ViT \cite{ebert2023plg}}          & \multicolumn{1}{c|}{34.8 M}                                 &  48.0                & -                 \\ \hline
\multicolumn{1}{l|}{Mask-RCNN \cite{he2017mask}}                & \multicolumn{1}{l|}{ResNet50 \cite{he2016deep}}        & \multicolumn{1}{c|}{43.7 M}                           & 38.2             & 34.7              \\
\multicolumn{1}{l|}{Mask-RCNN \cite{he2017mask} + ours}         & \multicolumn{1}{l|}{ResNet50 \cite{he2016deep}}        & \multicolumn{1}{c|}{45.9 M}                           & 39.6             & 35.9              \\
\multicolumn{1}{l|}{Mask-RCNN \cite{he2017mask} + ours}         & \multicolumn{1}{l|}{PLG-ViT \cite{ebert2023plg}}       & \multicolumn{1}{c|}{48.4 M}                                 & 45.4             & 41.4              \\ \hline
\end{tabular}
\end{table}

\begin{table}[t]
\scriptsize
\centering
\caption{\textbf{Comparison of detection and segmentation accuracy} on the LIVECell \cite{edlund2021livecell} test set. Different methods \cite{feng2021tood,he2017mask} are compared to our approach. We use TOOD \cite{feng2021tood} and Mask-RCNN \cite{he2017mask} as heads and ResNet50 \cite{he2016deep} as the backbone for all models.}
\label{tab:livecell}
\begin{tabular}{l|c|cc}
\hline
\multicolumn{1}{c|}{\multirow{2}{*}{\textbf{Method}}} & \multirow{2}{*}{\textbf{Params}} & \multicolumn{2}{c}{\textbf{Metrics}} \\
\multicolumn{1}{c|}{}                                 &                                  & \textbf{mAP\textsuperscript{bb}} & \textbf{mAP\textsuperscript{seg}} \\ \hline
TOOD \cite{feng2021tood}                                                  & 32.0 M                           & 29.4             & -                 \\
TOOD \cite{feng2021tood} + ours                                           & 32.8 M                           & 33.8             & -                 \\ \hline
Mask-RCNN \cite{he2017mask}                                            & 43.7 M                           & 36.8             & 37.3              \\
Mask-RCNN \cite{he2017mask} + ours                                      & 45.9 M                           & 38.0             & 38.0              \\ \hline
\end{tabular}
\end{table}

\begin{figure*}[t]
    % \centering
    %  \begin{subfigure}[b]{0.245\textwidth}
    %      \centering
    %      \includegraphics[width=1\textwidth]{images/comparison/faster_rcnn.png}
    %  \end{subfigure}
    %  \hfill
    %  \begin{subfigure}[b]{0.245\textwidth}
    %     \centering
    %      \includegraphics[width=1\textwidth]{images/comparison/tood.png}
    %  \end{subfigure}
    %  \hfill
    %  \begin{subfigure}[b]{0.245\textwidth}
    %      \centering
    %      \includegraphics[width=1\textwidth]{images/comparison/faster_rcnn_ours.png}
    %  \end{subfigure}
    % \hfill
    %  \begin{subfigure}[b]{0.245\textwidth}
    %      \centering
    %      \includegraphics[width=1\textwidth]{images/comparison/tood_ours.png}
    % \end{subfigure}
    % \hfill
    \centering
     \begin{subfigure}[b]{0.245\textwidth}
         \centering
         \includegraphics[width=1\textwidth]{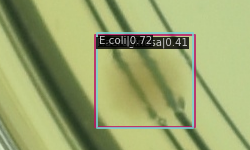}
     \end{subfigure}
     \hfill
     \begin{subfigure}[b]{0.245\textwidth}
        \centering
         \includegraphics[width=1\textwidth]{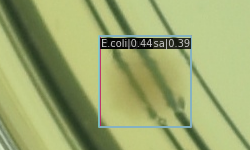}
     \end{subfigure}
     \hfill
     \begin{subfigure}[b]{0.245\textwidth}
         \centering
         \includegraphics[width=1\textwidth]{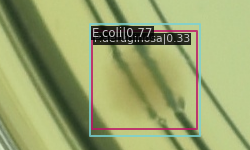}
     \end{subfigure}
    \hfill
     \begin{subfigure}[b]{0.245\textwidth}
         \centering
         \includegraphics[width=1\textwidth]{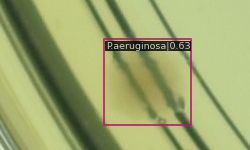}
    \end{subfigure}
    \hfill
    \begin{subfigure}[b]{0.245\textwidth}
         \centering
         \includegraphics[width=1\textwidth]{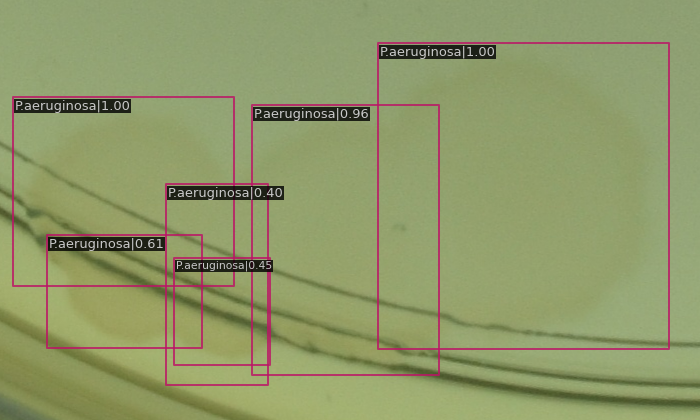}
         \caption{Faster-RCNN \cite{Ren_2017}}
         \label{fig:inf_rcnn}
     \end{subfigure}
     \hfill
     \begin{subfigure}[b]{0.245\textwidth}
        \centering
         \includegraphics[width=1\textwidth]{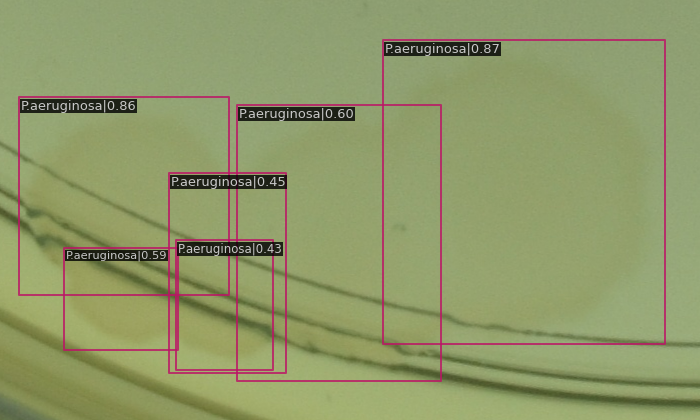}
         \caption{TOOD \cite{feng2021tood}}
         \label{fig:inf_TOOD}
     \end{subfigure}
     \hfill
     \begin{subfigure}[b]{0.245\textwidth}
         \centering
         \includegraphics[width=1\textwidth]{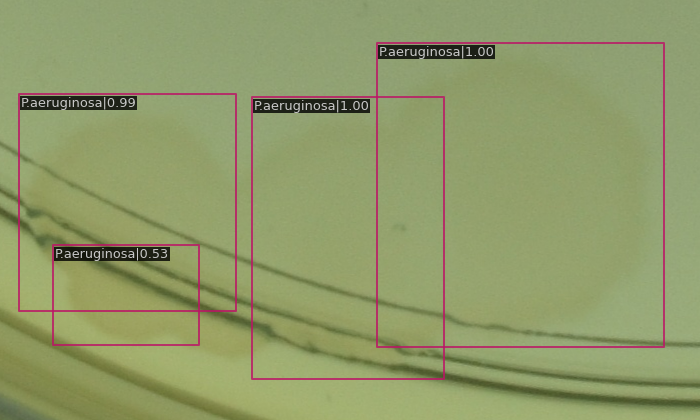}
         \caption{Faster-RCNN \cite{Ren_2017} + Ours}
         \label{fig:inf_rcnn_ours}
     \end{subfigure}
     \hfill
     \begin{subfigure}[b]{0.245\textwidth}
         \centering
         \includegraphics[width=1\textwidth]{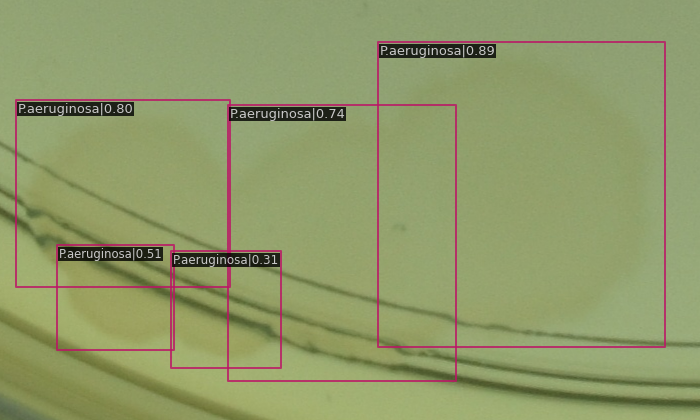}
         \caption{TOOD \cite{feng2021tood} + Ours}
         \label{fig:inf_TOOD_ours}
     \end{subfigure}
     
\caption{
    \textbf{Qualitative comparison} of Faster-RCNN \cite{Ren_2017} (a), TOOD \cite{feng2021tood} (b), Faster-RCNN + AttnPAFPN (c) and TOOD + AttnPAFPN (d). All networks are trained on the AGAR dataset \cite{majchrowska2021agar} under equal conditions. A small colony is visible in the first row with low contrast and distracting texture in the background. The second row shows a cluster of colonies with low contrast.
    }
    \label{fig:vergleich}
\end{figure*}

\begin{figure*}[t]
    \centering
     \begin{subfigure}[b]{0.33\textwidth}
         \centering
         \includegraphics[width=0.99\textwidth]{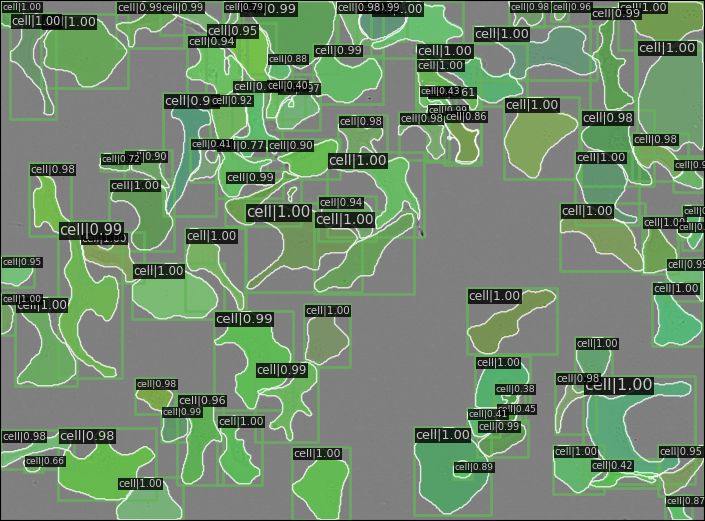}
     \end{subfigure}
     \hfill
    \begin{subfigure}[b]{0.33\textwidth}
         \centering
         \includegraphics[width=0.99\textwidth]{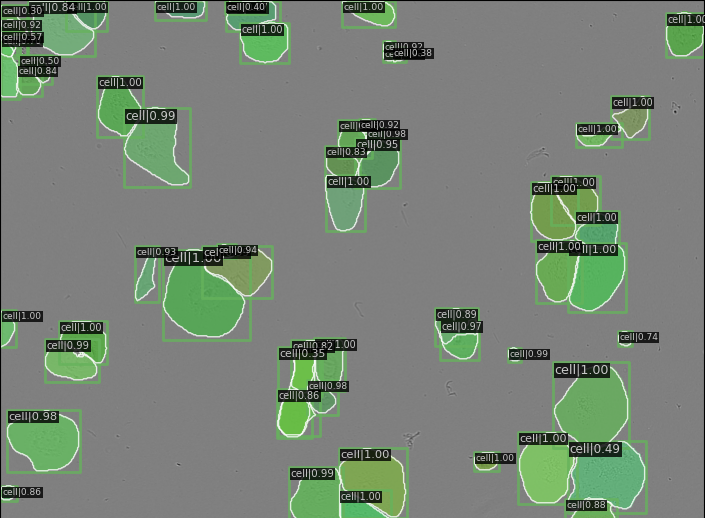}
     \end{subfigure}
     \hfill
    \begin{subfigure}[b]{0.33\textwidth}
         \centering
         \includegraphics[width=0.99\textwidth]{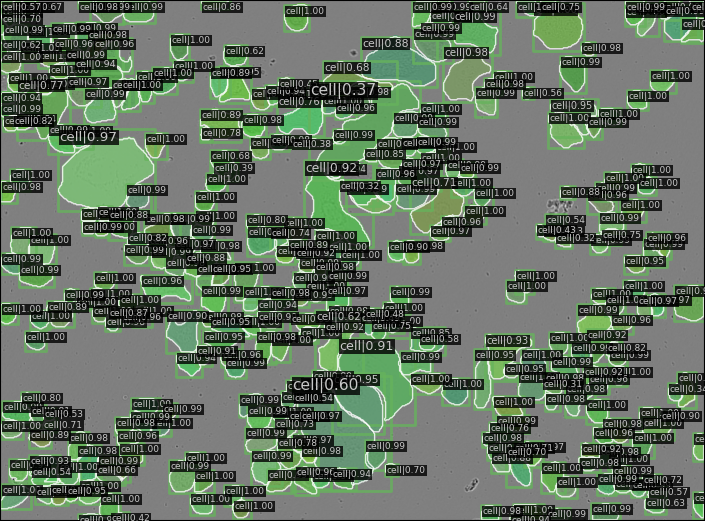}
     \end{subfigure}
     \hfill
\caption{
    \textbf{Qualitative result} of Mask-RCNN \cite{he2017mask} with our AttnPAFPN and ResNet50 \cite{he2016deep} on LIVECell \cite{edlund2021livecell}. 
    }
    \label{fig:livecell}
\end{figure*}

In addition to detecting bacteria colonies in high-resolution images, we also want to evaluate our method on medium-resolution images of other areas of application. 
For this purpose we use the COCO dataset \cite{lin2014microsoft}, which is a widespread baseline for object detection.% with about 118k images for training and 5k images for validation. 
For training the networks on COCO we use the standard settings \cite{chen2019mmdetection} proposed by the authors and train for 12 epochs. 
As network heads we use TOOD \cite{feng2021tood} and Mask-RCNN \cite{he2017mask} as an additional method for instance segmentation.

The results of the evaluation on COCO can be seen in Table \ref{tab:coco}. 
In contrast to AGAR, only a slightly improvement of the accuracy stemming from AttnPAFPN can be seen. 
As already noted in Section \ref{sec:rw}, this can be explained by the different characteristics of the COCO dataset, such as the relatively small number of tiny objects, in contrast to the AGAR dataset.
Using Mask-RCNN, on the other hand, the impact of our neck is more significant. 
We achieve $+1.4 / + 1.2 $ mAP for detection and segmentation, respectively.
The extension of the methods by a stronger transformer backbone increases the accuracy considerably.

We also performed experiments on the LIVECell dataset \cite{edlund2021livecell}, which is used to detect and segment cells in microscopy images. 
For this we also use TOOD and Mask-RCNN, which were previously pre-trained on COCO.
Additionally, we made some adjustments regarding the anchor-boxes of Mask-RCNN and TOOD as suggested by the authors of the dataset \cite{edlund2021livecell}. 
As a result, the networks are better adapted to the characteristics of the dataset.

The results on the LIVECell data set are listed in Table \ref{tab:livecell}. 
Here we can see that especially Mask-RCNN performs much better on the dataset than TOOD, which is a pure detection network. 
But especially TOOD benefits strongly from the extension by AttnPAFPN, which outperforms the baseline by $+3.4$ mAP. 
Mask-RCNN achieves with AttnPAFPN an increase in accuracy of $+1.2 / +0.7$ mAP for detection and segmentation, respectively. 
Figure \ref{fig:livecell} shows some visual results of our method with Mask-RCNN as head and ResNet50 as backbone.

\subsection{Visual Evaluation}
In addition to a quantitative evaluation we also present a qualitative evaluation on a greatly enlarged section of the image in Figure \ref{fig:vergleich}.
Here it can be seen that the conventional method has difficulties with particularly small and overlapping colonies, in contrast to our method.
In addition to the visual results on AGAR \cite{majchrowska2021agar}, typical results on the LIVECell dataset \cite{edlund2021livecell} can be seen in Figure \ref{fig:livecell}.

\section{Conclusion}
In this paper, we presented AttnPAFPN, a high-performance feature pyramid for high-resolution object detection.
Our AttnPAFPN uses our state-of-the-art efficient-global self-attention layers for better visual understanding.
Moreover, the efficient-global self-attention can be easily interchanged with any other self-attention mechanism.
Furthermore we add a additional scales to our PAFPN for predicting tiny and large objects on high- and low-resolution featuremaps, respectively.
In order to be executable even on resource-constrained hardware, we have considered efficiency and parameter count during the optimization of our method.
We have performed a comprehensive evaluation on a large scale public dataset \cite{zhang2021varifocalnet} for detecting bacterial colonies on agar dishes and proved the surpassing accuracy of our method compared to the current state-of-the-art.
In addition, we have performed experiments on the standard object detection baseline COCO \cite{lin2014microsoft}, as well as on LIVECell \cite{edlund2021livecell} for biomedical image analysis.

\section*{Acknowledgments} 
This work was supported by funding from the the Federal Ministry of Education and Research Germany in the project M\textsuperscript{2}Aind-DeepLearning (13FH8I08IA). Additional funding was provided by the German Research Foundation under grant number INST874/9-1 and by the Albert and Anneliese Konanz Foundation.

{\small
\bibliographystyle{ieee_fullname}
\bibliography{paper}

\begin{thebibliography}{10}\itemsep=-1pt

\bibitem{alexander1958automatic}
NE Alexander and DP Glick.
\newblock Automatic counting of bacterial cultures-a new machine.
\newblock {\em IRE Transactions on Medical Electronics}, 1958.

\bibitem{andreini2018deep}
Paolo Andreini, Simone Bonechi, Monica Bianchini, Alessandro Mecocci, and
  Franco Scarselli.
\newblock A deep learning approach to bacterial colony segmentation.
\newblock In {\em International Conference on Artificial Neural Networks
  (ICANN)}, 2018.

\bibitem{ba2016layer}
Jimmy~Lei Ba, Jamie~Ryan Kiros, and Geoffrey~E Hinton.
\newblock Layer normalization.
\newblock {\em arXiv preprint arXiv:1607.06450}, 2016.

\bibitem{beznik2022deep}
Thomas Beznik, Paul Smyth, Ga{\"e}l de Lannoy, and John~A Lee.
\newblock Deep learning to detect bacterial colonies for the production of
  vaccines.
\newblock {\em Neurocomputing}, 2022.

\bibitem{bochkovskiy2020yolov4}
Alexey Bochkovskiy, Chien-Yao Wang, and Hong-Yuan~Mark Liao.
\newblock Yolov4: Optimal speed and accuracy of object detection.
\newblock {\em arXiv preprint arXiv:2004.10934}, 2020.

\bibitem{cai2018cascade}
Zhaowei Cai and Nuno Vasconcelos.
\newblock Cascade r-cnn: Delving into high quality object detection.
\newblock In {\em Conference on Computer Vision and Pattern Recognition
  (CVPR)}, 2018.

\bibitem{detr}
Nicolas Carion, Francisco Massa, Gabriel Synnaeve, Nicolas Usunier, Alexander
  Kirillov, and Sergey Zagoruyko.
\newblock End-to-end object detection with transformers.
\newblock In {\em European Conference on Computer Vision (ECCV)}, 2020.

\bibitem{chen2019mmdetection}
Kai Chen, Jiaqi Wang, Jiangmiao Pang, Yuhang Cao, Yu Xiong, Xiaoxiao Li,
  Shuyang Sun, Wansen Feng, Ziwei Liu, Jiarui Xu, et~al.
\newblock Mmdetection: Open mmlab detection toolbox and benchmark.
\newblock {\em arXiv preprint arXiv:1906.07155}, 2019.

\bibitem{dosovitskiy2020image}
Alexey Dosovitskiy, Lucas Beyer, Alexander Kolesnikov, Dirk Weissenborn,
  Xiaohua Zhai, Thomas Unterthiner, Mostafa Dehghani, Matthias Minderer, Georg
  Heigold, Sylvain Gelly, et~al.
\newblock An image is worth 16x16 words: Transformers for image recognition at
  scale.
\newblock In {\em International Conference on Learning Representations (ICLR)},
  2020.

\bibitem{ebert2022multitask}
Nikolas Ebert, Patrick Mangat, and Oliver Wasenmuller.
\newblock Multitask network for joint object detection, semantic segmentation
  and human pose estimation in vehicle occupancy monitoring.
\newblock In {\em Intelligent Vehicles Symposium (IV)}, 2022.

\bibitem{ebert2023light}
Nikolas Ebert, Laurenz Reichardt, Didier Stricker, and Oliver Wasenm{\"u}ller.
\newblock Light-weight vision transformer with parallel local and global
  self-attention.
\newblock {\em arXiv preprint arXiv:2307.09120}, 2023.

\bibitem{ebert2023plg}
Nikolas Ebert, Didier Stricker, and Oliver Wasenm{\"u}ller.
\newblock Plg-vit: Vision transformer with parallel local and global
  self-attention.
\newblock {\em Sensors}, 2023.

\bibitem{edlund2021livecell}
Christoffer Edlund, Timothy~R Jackson, Nabeel Khalid, Nicola Bevan, Timothy
  Dale, Andreas Dengel, Sheraz Ahmed, Johan Trygg, and Rickard Sj{\"o}gren.
\newblock Livecell—a large-scale dataset for label-free live cell
  segmentation.
\newblock {\em Nature methods}, 2021.

\bibitem{falk2019u}
Thorsten Falk, Dominic Mai, Robert Bensch, {\"O}zg{\"u}n {\c{C}}i{\c{c}}ek,
  Ahmed Abdulkadir, Yassine Marrakchi, Anton B{\"o}hm, Jan Deubner, Zoe
  J{\"a}ckel, Katharina Seiwald, et~al.
\newblock U-net: deep learning for cell counting, detection, and morphometry.
\newblock {\em Nature methods}, 2019.

\bibitem{feng2021tood}
Chengjian Feng, Yujie Zhong, Yu Gao, Matthew~R Scott, and Weilin Huang.
\newblock Tood: Task-aligned one-stage object detection.
\newblock In {\em International Conference on Computer Vision (ICCV)}, 2021.

\bibitem{ferrari2017bacterial}
Alessandro Ferrari, Stefano Lombardi, and Alberto Signoroni.
\newblock Bacterial colony counting with convolutional neural networks in
  digital microbiology imaging.
\newblock {\em Pattern Recognition}, 2017.

\bibitem{geissmann2013opencfu}
Quentin Geissmann.
\newblock Opencfu, a new free and open-source software to count cell colonies
  and other circular objects.
\newblock {\em PloS one}, 2013.

\bibitem{gorokhov2022bacterial}
Oleg Gorokhov, Ramazan Fazylov, Maria Kazachuk, Ivan Lazukhin, Igor Mashechkin,
  Liudmila Pankratyeva, and Ivan Popov.
\newblock Bacterial colony detection method for microbiological photographic
  images.
\newblock In {\em International Joint Conference on Neural Networks (IJCNN)},
  2022.

\bibitem{he2017mask}
Kaiming He, Georgia Gkioxari, Piotr Doll{\'a}r, and Ross Girshick.
\newblock Mask r-cnn.
\newblock In {\em International Conference on Computer Vision (ICCV)}, 2017.

\bibitem{he2016deep}
Kaiming He, Xiangyu Zhang, Shaoqing Ren, and Jian Sun.
\newblock Deep residual learning for image recognition.
\newblock In {\em Conference on Computer Vision and Pattern Recognition
  (CVPR)}, 2016.

\bibitem{hendrycks2016gaussian}
Dan Hendrycks and Kevin Gimpel.
\newblock Gaussian error linear units (gelus).
\newblock {\em arXiv preprint arXiv:1606.08415}, 2016.

\bibitem{lamprecht2007cellprofiler}
Michael~R Lamprecht, David~M Sabatini, and Anne~E Carpenter.
\newblock Cellprofiler™: free, versatile software for automated biological
  image analysis.
\newblock {\em Biotechniques}, 2007.

\bibitem{lin2017feature}
Tsung-Yi Lin, Piotr Doll{\'a}r, Ross Girshick, Kaiming He, Bharath Hariharan,
  and Serge Belongie.
\newblock Feature pyramid networks for object detection.
\newblock In {\em Conference on Computer Vision and Pattern Recognition
  (CVPR)}, 2017.

\bibitem{lin2017focal}
Tsung-Yi Lin, Priya Goyal, Ross Girshick, Kaiming He, and Piotr Doll{\'a}r.
\newblock Focal loss for dense object detection.
\newblock In {\em International Conference on Computer Vision (ICCV)}, 2017.

\bibitem{lin2014microsoft}
Tsung-Yi Lin, Michael Maire, Serge Belongie, James Hays, Pietro Perona, Deva
  Ramanan, Piotr Doll{\'a}r, and C~Lawrence Zitnick.
\newblock Microsoft coco: Common objects in context.
\newblock In {\em European Conference on Computer Vision (ECCV)}, 2014.

\bibitem{liu2018path}
Shu Liu, Lu Qi, Haifang Qin, Jianping Shi, and Jiaya Jia.
\newblock Path aggregation network for instance segmentation.
\newblock In {\em Conference on Computer Vision and Pattern Recognition
  (CVPR)}, 2018.

\bibitem{liu2022two}
Shi-Jian Liu, Pin-Chao Huang, Xing-Sheng Liu, Jin-Jia Lin, and Zheng Zou.
\newblock A two-stage deep counting for bacterial colonies from multi-sources.
\newblock {\em Applied Soft Computing}, 2022.

\bibitem{liu2021swin}
Ze Liu, Yutong Lin, Yue Cao, Han Hu, Yixuan Wei, Zheng Zhang, Stephen Lin, and
  Baining Guo.
\newblock Swin transformer: Hierarchical vision transformer using shifted
  windows.
\newblock In {\em International Conference on Computer Vision (ICCV)}, 2021.

\bibitem{majchrowska2021agar}
Sylwia Majchrowska, Jaros{\l}aw Paw{\l}owski, Grzegorz Gu{\l}a, Tomasz Bonus,
  Agata Hanas, Adam Loch, Agnieszka Pawlak, Justyna Roszkowiak, Tomasz Golan,
  and Zuzanna Drulis-Kawa.
\newblock Agar a microbial colony dataset for deep learning detection.
\newblock {\em arXiv preprint arXiv:2108.01234}, 2021.

\bibitem{mansberg1957automatic}
HP Mansberg.
\newblock Automatic particle and bacterial colony counter.
\newblock {\em Science}, 1957.

\bibitem{naets2021mask}
Tanguy Naets, Maarten Huijsmans, Paul Smyth, Laurent Sorber, and Ga{\"e}l de
  Lannoy.
\newblock A mask r-cnn approach to counting bacterial colony forming units in
  pharmaceutical development.
\newblock {\em arXiv preprint arXiv:2103.05337}, 2021.

\bibitem{ramesh2018cell}
Nisha Ramesh and Tolga Tasdizen.
\newblock Cell segmentation using a similarity interface with a multi-task
  convolutional neural network.
\newblock {\em Journal of Biomedical and Health Informatics (JBHI)}, 2018.

\bibitem{Ren_2017}
Shaoqing Ren, Kaiming He, Ross Girshick, and Jian Sun.
\newblock Faster r-cnn: Towards real-time object detection with region proposal
  networks.
\newblock {\em Transactions on Pattern Analysis and Machine Intelligence
  (TPAMI)}, 2017.

\bibitem{rishav2021resfpn}
Rishav, René Schuster, Ramy Battrawy, Oliver Wasenmüller, and Didier
  Stricker.
\newblock Resfpn: Residual skip connections in multi-resolution feature pyramid
  networks for accurate dense pixel matching.
\newblock In {\em International Conference on Pattern Recognition (ICPR)},
  2021.

\bibitem{ronneberger2015u}
Olaf Ronneberger, Philipp Fischer, and Thomas Brox.
\newblock U-net: Convolutional networks for biomedical image segmentation.
\newblock In {\em International Conference on Medical image computing and
  computer-assisted intervention (MICCAI)}. Springer, 2015.

\bibitem{shamash2021onepetri}
Michael Shamash and Corinne~F Maurice.
\newblock Onepetri: accelerating common bacteriophage petri dish assays with
  computer vision.
\newblock {\em Phage}, 2(4):224--231, 2021.

\bibitem{srinivas2021bottleneck}
Aravind Srinivas, Tsung-Yi Lin, Niki Parmar, Jonathon Shlens, Pieter Abbeel,
  and Ashish Vaswani.
\newblock Bottleneck transformers for visual recognition.
\newblock In {\em Conference on Computer Vision and Pattern Recognition
  (CVPR)}, 2021.

\bibitem{tian2019fcos}
Zhi Tian, Chunhua Shen, Hao Chen, and Tong He.
\newblock Fcos: Fully convolutional one-stage object detection.
\newblock In {\em International Conference on Computer Vision (ICCV)}, 2019.

\bibitem{torelli2018autocellseg}
Angelo Torelli, Ivo Wolf, Norbert Gretz, et~al.
\newblock Autocellseg: robust automatic colony forming unit (cfu)/cell analysis
  using adaptive image segmentation and easy-to-use post-editing techniques.
\newblock {\em Scientific reports}, 2018.

\bibitem{vaswani2017attention}
Ashish Vaswani, Noam Shazeer, Niki Parmar, Jakob Uszkoreit, Llion Jones,
  Aidan~N Gomez, {\L}ukasz Kaiser, and Illia Polosukhin.
\newblock Attention is all you need.
\newblock {\em Neural Information Processing Systems (NeurIPS)}, 2017.

\bibitem{wang2020cspnet}
Chien-Yao Wang, Hong-Yuan~Mark Liao, Yueh-Hua Wu, Ping-Yang Chen, Jun-Wei
  Hsieh, and I-Hau Yeh.
\newblock Cspnet: A new backbone that can enhance learning capability of cnn.
\newblock In {\em Conference on Computer Vision and Pattern Recognition
  Workshops (CVPR Workshops)}, 2020.

\bibitem{wang2022pvt}
Wenhai Wang, Enze Xie, Xiang Li, Deng-Ping Fan, Kaitao Song, Ding Liang, Tong
  Lu, Ping Luo, and Ling Shao.
\newblock Pvt v2: Improved baselines with pyramid vision transformer.
\newblock {\em Computational Visual Media}, 2022.

\bibitem{xie2021segformer}
Enze Xie, Wenhai Wang, Zhiding Yu, Anima Anandkumar, Jose~M Alvarez, and Ping
  Luo.
\newblock Segformer: Simple and efficient design for semantic segmentation with
  transformers.
\newblock {\em Advances in Neural Information Processing Systems (NeurIPS)},
  2021.

\bibitem{zhang2021varifocalnet}
Haoyang Zhang, Ying Wang, Feras Dayoub, and Niko Sunderhauf.
\newblock Varifocalnet: An iou-aware dense object detector.
\newblock In {\em Conference on Computer Vision and Pattern Recognition
  (CVPR)}, 2021.

\bibitem{zhu2019deformable}
Xizhou Zhu, Han Hu, Stephen Lin, and Jifeng Dai.
\newblock Deformable convnets v2: More deformable, better results.
\newblock In {\em Conference on Computer Vision and Pattern Recognition}, 2019.

\bibitem{zhu2020deformable}
Xizhou Zhu, Weijie Su, Lewei Lu, Bin Li, Xiaogang Wang, and Jifeng Dai.
\newblock Deformable detr: Deformable transformers for end-to-end object
  detection.
\newblock In {\em International Conference on Learning Representations (ICLR)},
  2021.

\end{thebibliography}
}

\end{document}